\documentclass[10pt, conference, compsocconf]{IEEEtran}
\ifCLASSINFOpdf
\else
\fi
\usepackage[utf8]{inputenc}
\usepackage{gensymb}
\usepackage{graphicx}
\usepackage{subcaption}
\graphicspath{ {./images/} }
\usepackage{hyperref}
\usepackage{amssymb}
\usepackage{mathtools}

\begin{document}
%



\title{Dynamic Deep Multi-task Learning for Caricature-Visual Face Recognition}
\author{Zuheng Ming, \quad Jean-Christophe Burie, \quad Muhammad Muzzamil Luqman\\
    L3i, La Rochelle University, France\\
    \{zuheng.ming, jcburie, mluqma01\}@univ-lr.fr}

\maketitle

\begin{abstract}
Rather than the visual images, the face recognition of the caricatures is far from the performance of the visual images. The challenge is the extreme non-rigid distortions of the caricatures introduced by exaggerating the facial features to strengthen the characters. In this paper, we propose dynamic multi-task learning based on deep CNNs for cross-modal caricature-visual face recognition. Instead of the conventional multi-task learning with fixed weights of the tasks, the proposed dynamic multi-task learning dynamically updates the weights of tasks according to the importance of the tasks, which enables the training of the networks focus on the hard task instead of being stuck in the overtraining of the easy task. The experimental results demonstrate the effectiveness of the dynamic multi-task learning for caricature-visual face recognition. The performance evaluated on the datasets CaVI and WebCaricature show the superiority over the state-of-art methods. The implementation code is available here.
\footnote{\url{https://github.com/hengxyz/cari-visual-recognition-via-multitask-learning.git}}

\end{abstract}


\IEEEpeerreviewmaketitle

\section{\textbf{Introduction}}
\begin{figure}[t]
\begin{center}
   \includegraphics[width=0.9\linewidth]{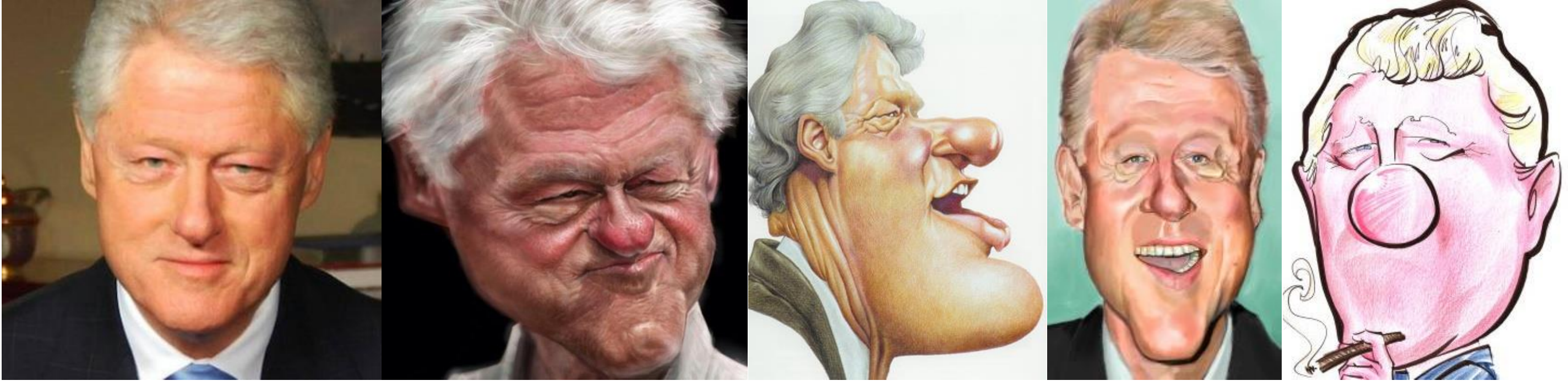}
\end{center}
   \caption{Realistic visual images and the caricatures of Bill Clinton. The different artistic styles result the large variation of the caricatures even with the same identity.}
\label{fig:caricatures}
\end{figure}

\begin{figure*}[t]
\begin{center}
   \includegraphics[width=0.9\linewidth]{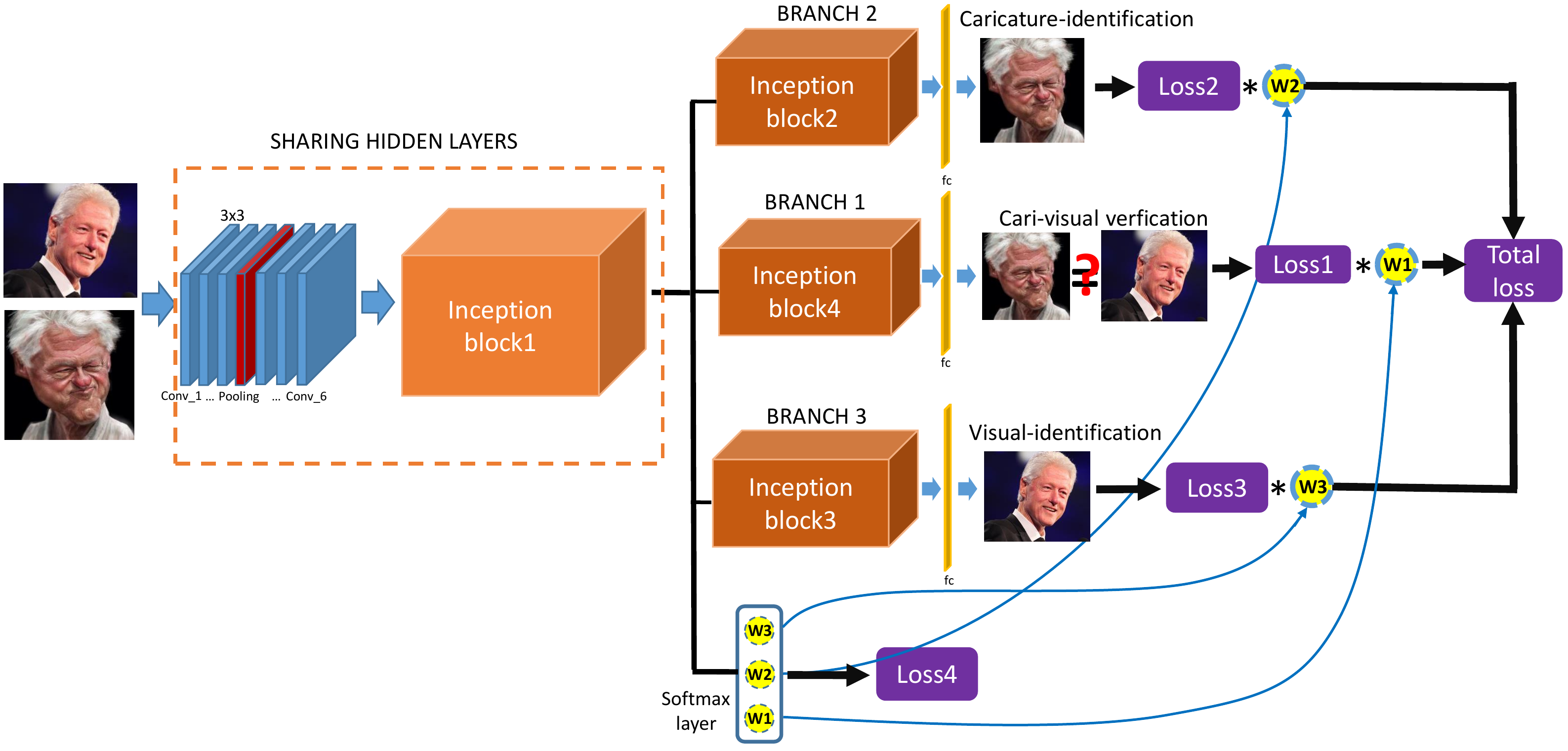}
\end{center}
   \caption{The proposed multi-task learning framework with dynamic weights of tasks for cross-modal caricature-visual face recognition. Different recognition modalities are learned by the different tasks. The introduced dynamic weights module can update the weights of tasks according to the difficulty of the training of tasks.}
\label{fig:framework}
\end{figure*}

In the recent decade, the performance of face recognition ~\cite{taigman2014deepface, schroff2015facenet, liu2017sphereface} achieves or surpasses human being performance on the datasets such as LFW\cite{huang2007labeled}, YTF\cite{wolf2011face} etc. Rather than the methods based on the hand-craft features such as LBP, Gabor-LBP, HOG, SIFT~\cite{ahonen2006face}, the deep learning based methods mitigate the problems, e.g. the occlusion, the variation of the pose, via the representation learning by leveraging the enormous data.  Nonetheless, the challenge of face recognition still exists, for example, the non-rigid deformation and distortion as shown in the facial expression as well as in the caricatures. Unlike the realistic  visual facial image, caricatures are the facial artistic drawings with the exaggerations to strengthen certain facial instinct features as shown in ~\figurename~\ref{fig:caricatures}.  Due to the diverse artistic styles, the caricatures not only have great difference with the heterogeneous real visual images but also differ greatly within the caricatures. Either the intra-class or the inter-class variation of caricatures are much more distinct than the visual images~\cite{HuoBMVC2018WebCaricature}, which is a big challenge comparing to the face recogntion on the real visual images. 
Thus it is not plausible to employ a model trained for the real visual image to recognize the caricatures and vice versa. 
Rather than the single-task method such as ~\cite{HuoBMVC2018WebCaricature}, the caricature-visual recognition cross two modalities ideally suit the multi-task learning which can learn the two different recognition modalities by two different specific tasks integrated in a unified networks. Despite the face recognition on caricatures and visual images cross different modalities, they still share some common intrinsic features of face. This is our motivation to propose to use the hard parameter sharing structure for multi-task learning~\cite{ruder2017overview} rather than the Siamese couple networks~\cite{garg2018deep} in our work, in which the sharing hidden layers used for learning common latent features are shared between all tasks.
The multi-task learning is substantially an optimization problem for multiple objectives. While the different tasks may have different importances and also have the different training difficulty , thus how to find the optimal weights of the tasks is an important issue in the multi-task learning. Some works simply assign the equal weights to the  different tasks as described in~\cite{chen2017multi}. However many works prove that  the performance varies in function of the weights in the multi-task learning and the optimal performance can be obtained by the weighted task with different weights~\cite{kendall2018multi}.
 There are two ways to search the optimal weights for the deep multi-task learning: 1) the static method; 2) the dynamic method. In the static method, the weights of tasks are searched either by experimental methods such as~\cite{garg2018deep} or by a greedy search~\cite{tian2015pedestrian}. The weights of tasks are fixed during the training of the weights of networks. Instead of exhausting and time consuming manual searching methods, the dynamic methods are capable to update the weights of tasks automatically according to the variation of the networks such as the variation of the loss or the gradients~\cite{ kendall2018multi, zhang2016learning, chen2017gradnorm, yin2018multi}. 
 In this work, we propose a dynamic multi-task learning method based on the deep CNNs to employ the caricature-visual face recognition. The proposed method can dynamically generate the weights of tasks according to the current importance of the tasks during the training. Rather than the existed methods, our method conducts the training of the networks focus on the hard task instead of being stuck in the overtraining of the easy task which enable the training more efficient. The proposed multi-task learning framework models the three different recognition tasks, i.e. caricature recognition, visual image recognition and caricature-visual face verification, by three different branches based on the sharing hidden layers. And the dynamic weights of the tasks are generated by a softmax layer connecting to the last layer of the hidden sharing layers (see \figurename~\ref{fig:framework}). Unlike~\cite{kendall2018multi, zhang2016learning, chen2017gradnorm}, no more hyperparameter is introduced in our method for updating the dynamic weights. 

In a summary, our main contributions of this paper are detail below.
\begin{itemize}
	\item We propose a dynamic deep multi-task learning approach for the cross-modal caricature-visual face recognition, which enable the multi-task learning more efficiently by focusing on the training of the hard task instead of overtraining of the easy task. 
    \item Both the theoretical analysis and the experimental results demonstrate the effectiveness of the proposed dynamic weights for helping the training.
    \item We have demonstrated that, for all the three recognition modalities,  the proposed multi-task learning can outperform the state-of-the-art performance on the datasets CaVI and WebCaricature.
\end{itemize}

The remainder of this paper is organized as follows: Section II briefly reviews the related works; Section III describes the architecture of the dynamic multi-task network proposed in this work. Section IV presents the approach of multi-task learning with dynamic weights following by Section V where the experimental results are analyzed. Finally, in Section VI, we draw the conclusions and present the future works.


\section{\textbf{Related Work}}
\textbf{Caricature-Visual face recognition} By virtue of the deep neural networks especially the deep CNNs, face recognition has made a series of breakthrough in the recent decade, such as DeepFace~\cite{taigman2014deepface}, DeepID~\cite{sun2015deeply} series, FaceNet~\cite{schroff2015facenet}, VGG face~\cite{simonyan2014very} have all outperform the human being on LFW (97.35\%-99.63\%) and YTF (91.4\%-95.12\%). Recently SphereFace~\cite{liu2017sphereface} achieves the state-art-art performance on dataset MegaFace~\cite{kemelmacher2016megaface}.
However, due to the challenge of the cross-modal heterogeneous face matching problem and the lack of the dataset, the caricature-visual face recognition is not sufficiently studied especially with the deep learning based methods. Huo and al.~\cite{HuoBMVC2018WebCaricature} propose a large caricature dataset WebCaricature consisting of 252 people with 6024 caricatures and 5974 photos. It shows that the performance of the deep learning based method with pretrained VGG-Face is significantly better than the hand-craft feature based methods such as SIFT,  Gabor etc, i.e. 57.22\% for caricature-visual face verification,  and 55.41\%@Rank-1 accuracy for caricature to real visual image identification while 55.53\%@Rank-1 accuracy for real visual image to caricature identification.  However, the performance of the proposed method is still limited. Garg et al.~\cite{garg2018deep} propose a CNN-based coupled-networks CaVINet consisting 13 convolutional layers of VGGFace for caricature-verification and caricature recognition. This work also introduce a new publicly available dataset (CaVI) that contains 5091 caricatures and 6427 visual images. The CaVINet achieves 91.06\% accuracy for the caricature-visual face verification task, 85.09\% accuracy for caricature identification task and 94.50\% accuracy for caricature identification task. However, the weight of tasks are manually searched by the experimental method. 

\textbf{Multi-task learning} has been used successfully to computer vision. Fast R-CNN~\cite{girshick2015fast} uses a multi-task loss to jointly train the classification and bounding-box regression for object detection. The classification task is set as the main task with the weight 1 and the bounding-box regression is set as the side task weighted by $\lambda$. It shows the multi-task learning is superior to the single-task learning for object detection. Tian et al.~\cite{tian2015pedestrian} fix the weight for the main task to 1, and obtain the weights of all side tasks via a greedy search within 0 and 1. In ~\cite{chen2017gradnorm} the weights is updated dynamically by the loss of the gradients meanwhile an hyperparameter is introduced for balancing the training of different tasks. ~\cite{kendall2018multi} introduces a uncertainty coefficient $\theta$ to revise the loss function which can be fixed manually or learned based on the total loss.   Zhang et al.~\cite{zhang2016learning} introduce an hyperparameter $\rho$ as a scale factor to calculate the dynamic weight $\lambda_t$ of face attributes recognition.  Yin et al.~\cite{yin2018multi} proposed a multi-task model for face pose-invariant recognition. The weights of tasks are updated by the total loss of networks leading the training to be stuck in the easy task.

\section{Architecture}
The proposed networks for multi-task learning is based on the hard parameter sharing structure (see ~\figurename~\ref{fig:framework}), in which the sharing hidden layers can capture the modality-common features between all tasks 
such as the face-like pattern, the similar topological structure of the eyes, nose, mouth etc. 
The task-specific branches to learn the modality-specific features, i.e. caricature recognition, face recognition and caricature-visual face verification. 
The three branches have almost identical structures to facilitate the transfer learning from the pretrained face recognition task. Specifically, BRANCH 1 can extract the embedded features of bottleneck layer for caricature-visual face verification, and BRANCH 2 and 3 use the fully connected softmax layer to calculate the probabilities of the identities of the input caricatures or real visual images. The deep CNNs with the Inception-ResNet structure have overall 13 million parameters of about 20 hidden layers in terms of the depth, whose parameters are much fewer than VGGFace with ~138 million parameters.The dynamic weights of tasks are generated by the softmax layer connecting to the end of the sharing hidden layers, which can be so called the dynamic-weight-generating-module.Each element in the dynamic-weight-generating-module is corresponding to a weight of a task $w_i$.


\section{Dynamic Deep Multi-task learning}
The total loss of the proposed multi-task CNNs is the sum of the weighted losses of the multiple tasks.

(I) \noindent~\textbf{Multi-task loss $\mathcal{L}$}: The multi-task total loss $\mathcal{L}$ is defined as follows:
\begin{equation}
\label{eq1}
\mathcal{L}(\mathbf{X};\Theta;\Psi) = \sum_{i=1}^{T} w_i(\Psi)\mathcal{L}_i(\mathbf{X}_i;\Theta_i)
\end{equation}
where $T$ is the number of the tasks, here $T=3$. ${X_i}$ and ${\Theta_i}$ are the feature and the parameters corresponding to each task $i$, $\Theta=\{\Theta_i\}_{i=1}^{T}$ are the overall parameters of the networks to be optimized by the total loss $\mathcal{L}$.   
The parameters of the softmax layer in the dynamic-weight-generating-module is denoted as $\Psi$ which is used to generate the dynamic weights  $w_i \in[0,1]$ s.t. $\sum w_i=1$. Note that the $\Psi \not\in \Theta$.  
Thus \{$\mathbf{X_i}$, $\Theta_i$\}$\in \mathbb{R}^{d_i}$
where $d_i$ is the dimension of the features $X_i$, and $\{\mathcal{L}_i, w_i\}\in \mathbb{R}^1$.  Particularly, when $w_i$ = 1 and $w_{j\neq i}$ = 0 the multi-task networks are degraded as the single-task networks. 
For  example, $w_1$ = 1 and $w_2$=0, $w_3$=0, is degraded to the single task network for caricature recognition (i.e. consisting of BRANCH 1 and the sharing hidden layers). 

(II)\noindent~\textbf{Caricature-Visual face verification task loss} $\mathcal{L}_1$: The loss for caricature-visual face verification task is measured by the center loss~\cite{wen2016discriminative} joint with the cross-entropy loss of softmax of BRANCH 1. The loss function $\mathcal{L}_1$ is given by:      
\begin{equation}
\label{L1}
\mathcal{L}_1(\mathbf{X}_1;\Theta_1) = \mathcal{L}_{s1}(\mathbf{X}_1;\Theta_1) + \alpha\mathcal{L}_c(\mathbf{X}_1;\Theta_1)
\end{equation}

where $\mathcal{L}_{s1}$ is the cross-entropy loss of softmax of BRANCH 1, $\mathcal{L}_{c}$ is the center loss weighted by the hyperparameter $\alpha$. The $\mathcal{L}_{c}$ can be treated as a regularization item of softmax loss $\mathcal{L}_{s1}$ which is given by:
\begin{equation}
\label{eq:cross_entropy} 
\mathcal{L}_{s1}(\mathbf{X}_1;\Theta_1)= \sum^{K}_{k=1}-y_klogP(y_k=1|\mathbf{X}_1,\theta_k) 
\end{equation}

where $K$ is the number of  identities in the training dataset, ${y_k}\in\{0,1\}$ is the label of the feature $\mathbf{X}_1$, $P(y_k|\mathbf{X}_1,\theta_k)$ is softmax function. The bottleneck layer of BRANCH 1 is extracted as the feature $\mathbf{X}_1$ of the input image. The center loss $\mathcal{L}_c$ is given by:

\begin{equation}
\label{eq:centerloss} 
\mathcal{L}_c(\mathbf{X}_1;\Theta_1) = ||\mathbf{X}_1-C_{y_k}||
\end{equation}
Where the $C_{y_k}$ is the center of the class which $\mathbf{X}_1 $ belonging to, $C_{y_k} \in \mathbb{R}^{d_1}$.

(III)\noindent~\textbf{Caricature identification task loss $\mathcal{L}_2$, and Visual identification task loss $\mathcal{L}_3$} : The loss function $\mathcal{L}_2$ and $\mathcal{L}_3$ are the cross-entropy loss of the softmax layer of BRANCH 2 and BRANCH 3 respectively. The equations of $\mathcal{L}_2$, $\mathcal{L}_3$ are as same as Equation~\ref{eq:cross_entropy}, and the $K$ in $\mathcal{L}_{2}$ or $\mathcal{L}_{3}$ is the number of the identities, $\mathbf{X}_2$ or $\mathbf{X}_3$ is the bottleneck layer of BRANCH 2 or BRANCH 3. 


(IV)\noindent~\textbf{Generation of the dynamic weights $w_i(\Psi)$}:
The dynamic weights $w_i$ are generated by the softmax layer of the dynamic-weight-generating-module which is given by: 
\begin{equation}
\label{eq1_dynamicweights}
w_i(\mathbf{Z};\Psi) = \frac{e^{f^{\psi_i}(\mathbf{Z})}}{\sum^{T}_{i'}e^{f^{\psi_{i'}}(\mathbf{Z})}}
\end{equation}
where the $\mathbf{Z} \in \mathbb{R}^{d_z}$ is the flattened output of the last layer of the sharing hidden layers. $T$ is the number of the tasks, here $T$=3. $\psi_i$ is parameters in the softmax layer of the dynamic-weight-generating-module $\{\psi_i\}_{i=1}^{T}=\Psi$, $\psi_i\in \mathbb{R}^{d_z}$. $f^{\psi_{i}}(\mathbf{Z})$ is activation function which is given by:
\begin{equation}
\label{activefunction}
f^{\psi_i}(\mathbf{Z}) = \psi_i\mathbf{Z}^T + b_i
\end{equation}
Note that, we do not use the Relu function as the activation function since Relu discards the values minors zero. This shrinks the range of the variation of the dynamic weights $w_i$.
 
(V) \noindent~\textbf{Update of the dynamic weights $w_i$}:
We propose a new loss function to update the dynamic weights which can drive the networks always train the hard task:
\begin{equation}
\label{eq:L3} 
\mathcal{L}_{4}(\mathbf{Z};\Psi)=\sum^{T}_{i=1}\frac{w_i(\psi_i)}{\mathcal{L}_i(\Theta_i)} \quad s.t. \quad \sum w_i=1
\end{equation}
Note that, $\mathcal{L}_i\{\Theta_i\}$ is independent with $w_i(\psi_i)$ since $\Theta_i\cap \psi_i=\emptyset$ , $i \in [1,..,T]$, thus $\mathcal{L}_i$ is constant for the dynamic weight update loss function $\mathcal{L}_4$.
 
(VI)\noindent~\textbf{Analysis of the dynamic weights} Here we show how the proposed dynamic weights drive the networks focus on training the hard task. Considering the Equation~\ref{eq1_dynamicweights} and Equation~\ref{activefunction}, the gradient of the $\psi_i$ can be given by
\begin{equation}
\label{eq:gradientofweights} 
\nabla{\psi_i}=\frac{\partial \mathcal{L}_4}{\partial \psi_i} =\frac{1}{\mathcal{L}_i}\frac{\partial w_i(\psi_i)}{\partial \psi_i}= \frac{1}{\mathcal{L}_i}\frac{a_i\sum^{T}_{j\neq i}a_{j}}{(\sum_i^T a_i)^2}\mathbf{Z}
\end{equation}
where $a_i = e^{{\psi_i}\mathbf{Z}^T+b_i}$, and the update of the parameters is  $\psi_i^{t+1}=\psi_i^{t}-\eta\nabla{\psi_i}^t$ where $\eta$ is the learning rate. Then the new value of the dynamic weight $w_i^{t+1}$ can be obtained by the Equation~\ref{eq1_dynamicweights} with the  $\psi_i^{t+1}$. 
We assume the $b_i^0=0, \psi_i^0=0, \eta=1$,  (this is possible if we initialize the $\psi_i, b_i$ by zero), the $\psi_i^t$ can be given by
\begin{equation}
\label{eq:parametersupdate} 
\psi_i^t=-\sum\frac{1}{\mathcal{L}_i}\frac{a_i\sum^{T}_{j\neq i}a_{j}}{(\sum_i^T a_i)^2}\mathbf{Z}
\end{equation}
if we consider the case for two tasks $w_1$ and $w_2$:
\begin{equation}
\label{eq:dynamicweightsanaylsis} 
\begin{split}
\frac{w_1^t}{w_2^t}
&=e^{(\psi_1^t-\psi_2^t)\mathbf{Z}^T} \\
&=e^{(\frac{1}{\mathcal{L}_2}-\frac{1}{\mathcal{L}_1})\frac{a_1a_2}{(a_1+a_2)^2}\mathbf{Z}\mathbf{Z}^T}
\end{split}
\end{equation}
We can see that $a_i>0$ and $ZZ^T\ge 0$, so if $\mathcal{L}_2 < \mathcal{L}_1$ the $\frac{w_1}{w_2}>1$ namely $w_1>w_2$. It means if the loss of task1 larger than the loss of task 2, the weight of the task1 is larger than the one of task2. It indicates that the proposed loss function $\mathcal{L}_3$ can well update the weights of tasks to drive the networks always train the hard task firstly.

(VII)\noindent~\textbf{Training protocol}:  The training of the entire deep CNNs includes two independent training: the training of the parameters of the networks {$\Theta$} by the multi-task loss $\mathcal{L}(\Theta)=\sum_{i=1}^{3}\mathcal{L}_i(\theta_i)$ and the training of the parameters of weight-generate-module {$\Psi$} by the loss $\mathcal{L}_4(\Psi)$. These  can be conducted simultaneously in a parallel way.
\begin{equation}
\Theta^{t-1}-\eta\frac{\partial\mathcal{L}(\Theta)}{\partial\Theta}\mapsto\Theta^{t}
\end{equation}
\begin{equation}
\Psi^{t-1}-\eta\frac{\partial\mathcal{L}_4(\Psi)}{\partial\Psi}\mapsto \Psi^{t}
\end{equation}

\section{Experiments and analysis}
\subsection{Datasets}
CaVI and WebCaricature are so far the most large public datasets for caricature-visual recognition research. In this work, both of the datasets are used to train and evaluate our proposed model.


\subsection{Pretrained Model}
Since either the dataset CaVI and WebCaricature is relative small to train ta deep CNNs for face recognition problems, before the training of the proposed multi-task CNNs, a single-task network constituted of the sharing hidden layers and the BRANCH 3 is pretrained for face verification-task with large-scale dataset MSCeleb-1M~\cite{guo2016ms}. MTCNN~\cite{zhang2016joint} is used for obtaining the images of faces from the raw images. The RMSprop with the mini-batches of 90 samples are applied for optimizing the parameters. The momentum coefficient is set to 0.99. The learning rate is started from 0.1, and divided by 10 at the 60K, 80K iterations respectively. The dropout probability is set 0.5 and the weight decay is 5e-5. The networks are initialized by Xavier with the zero bias. Then the training of the dynamic multi-task CNNs can handling on the pretrained model such as initialized the BRANCH2 and BRANCH3 by the trained BRANCH1.

\begin{figure*}
\centering
\begin{minipage}{.24\textwidth}
  \centering
  \includegraphics[width=\linewidth]{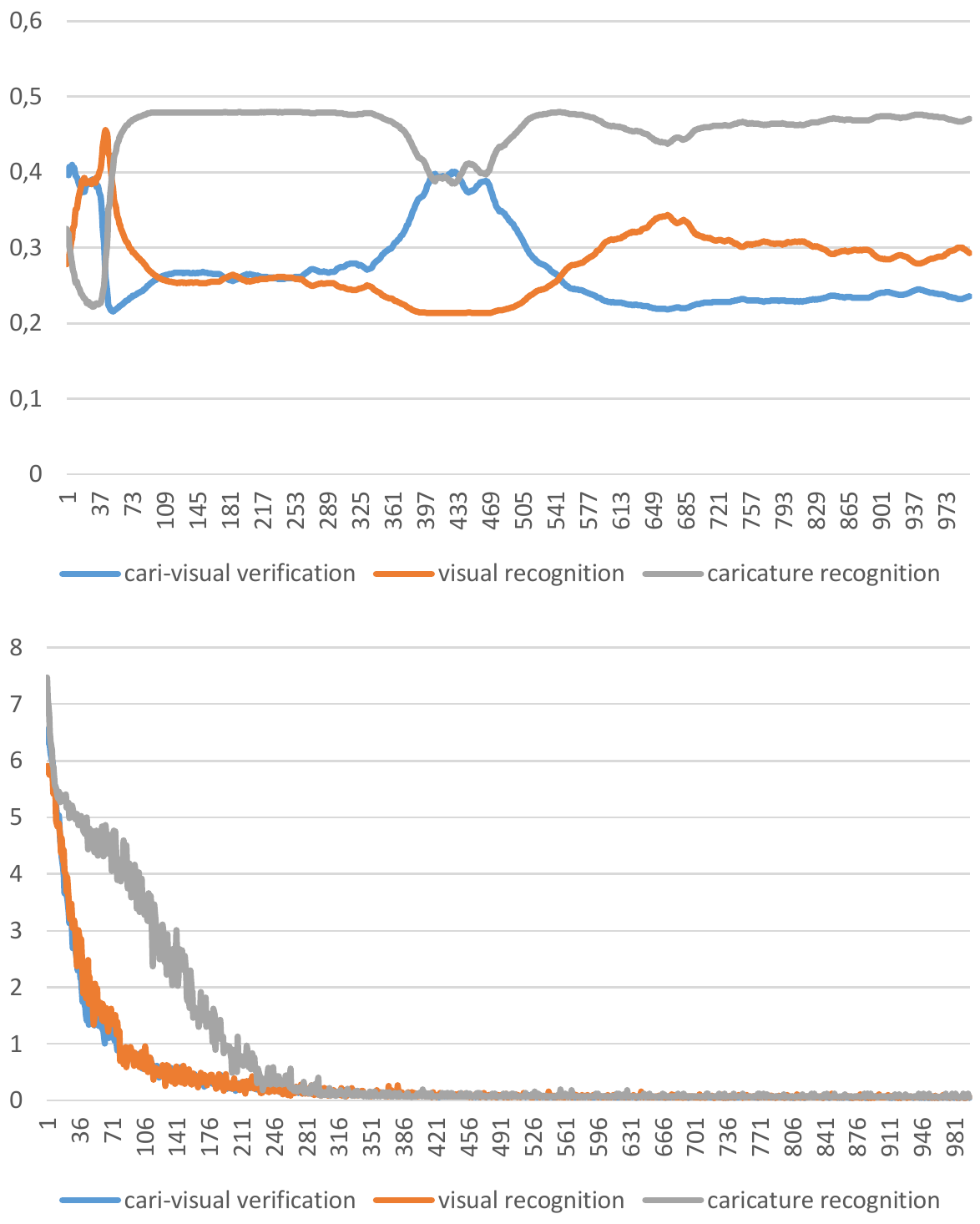}
  \subcaption{Our Dynamic-CAVI}
  \label{fig:dynweights_real_CK+}
\end{minipage}%
\begin{minipage}{.24\textwidth}
  \centering
  \includegraphics[width=\linewidth]{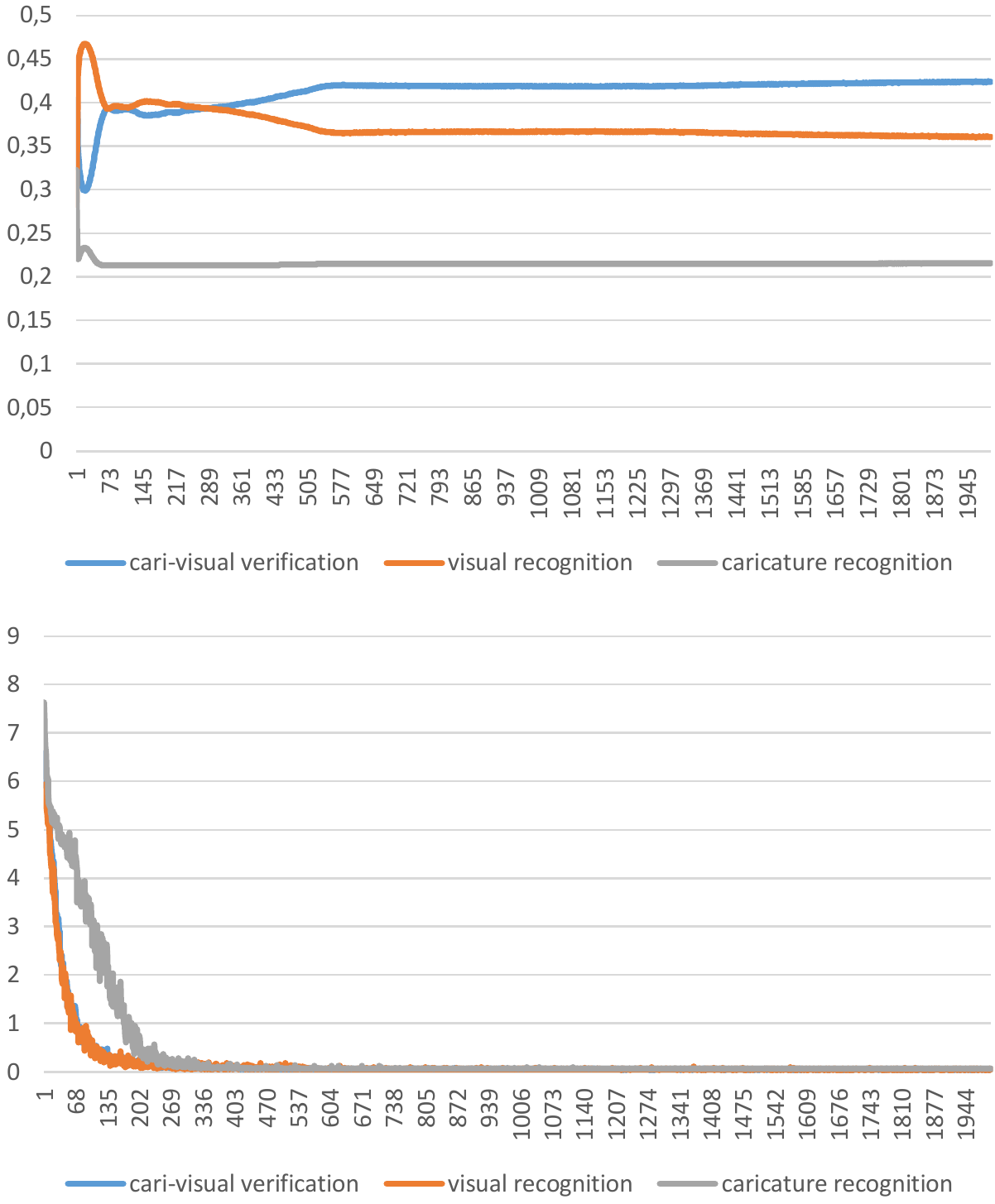}
  \subcaption{Naive Dynamic-CAVI}
  \label{fig:dynweights_real_oulu}
\end{minipage}
\begin{minipage}{.25\textwidth}
  \centering
  \includegraphics[width=\linewidth]{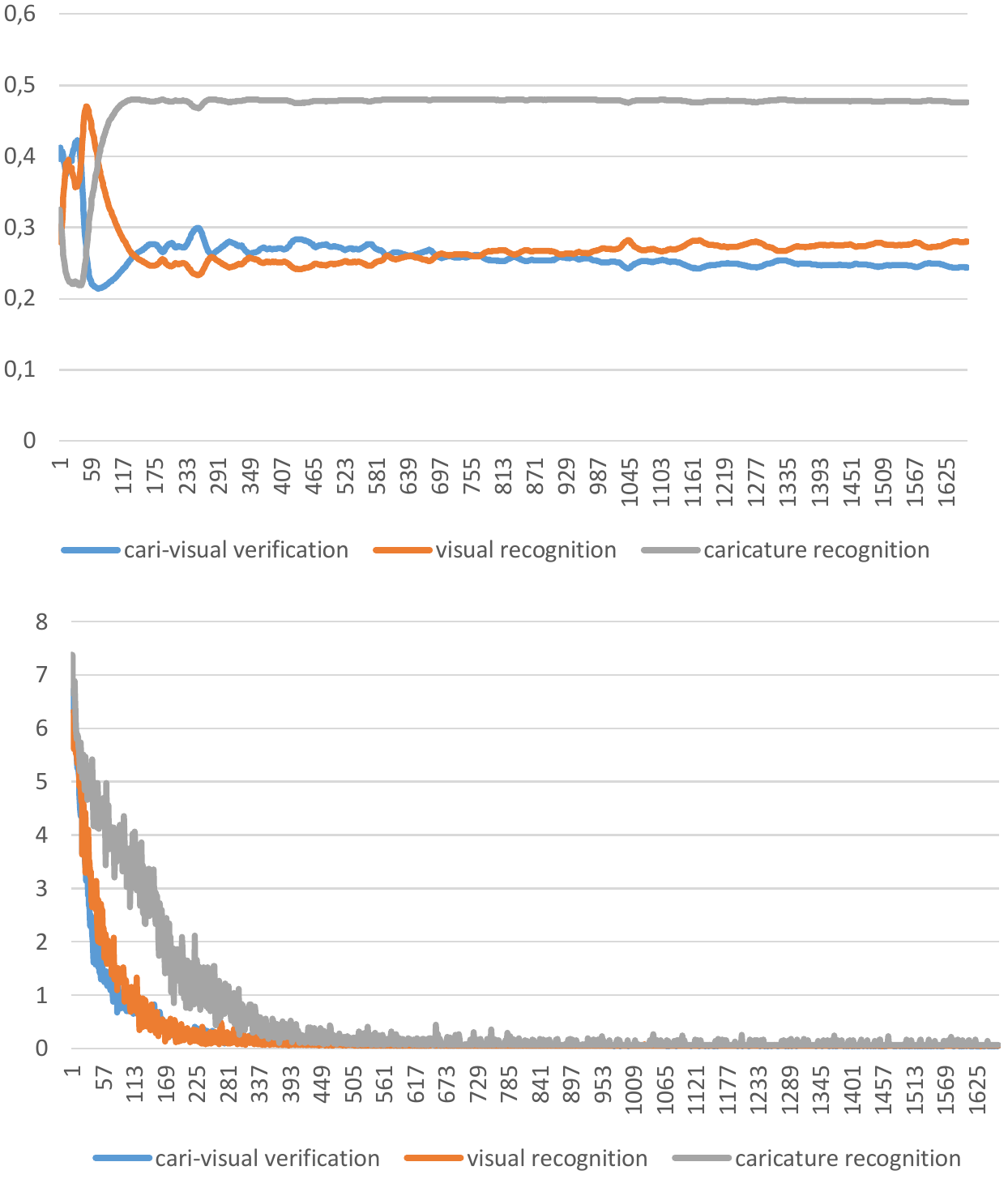}
  \subcaption{Our Dynamic-WebCari}
  \label{fig:dynweights_fake_CK+}
\end{minipage}
\begin{minipage}{.25\textwidth}
  \centering
  \includegraphics[width=\linewidth]{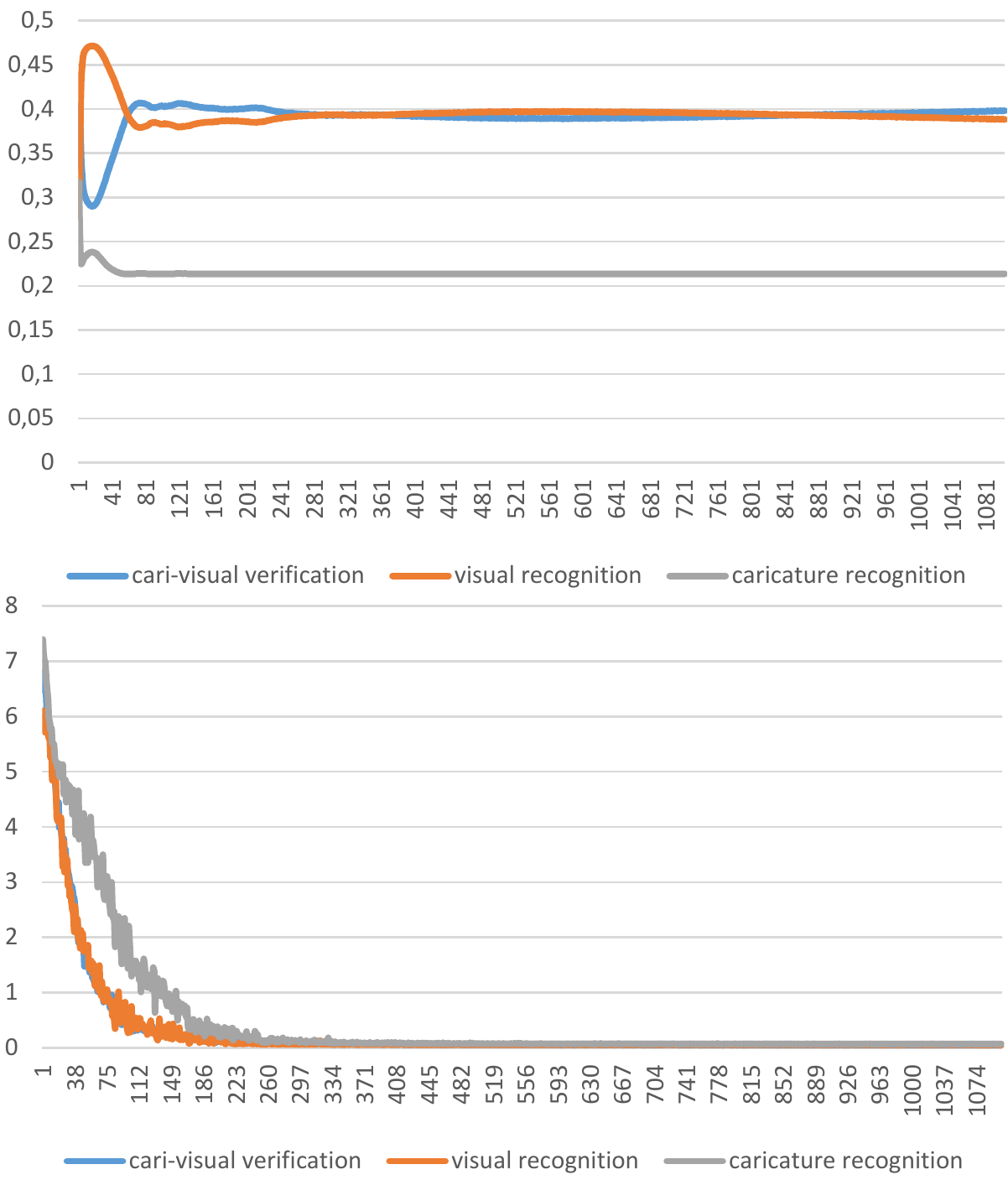}
  \subcaption{Naive Dynamic-WebCari}
  \label{fig:dynweights_fake_oulu}
\end{minipage}
\caption{The upper row shows the dynamic weights for both methods and the bottom row shows the corresponding losses.  The vertical axis is the identification accuracy and the horizontal axis is the number of the training iterations. Curves on different colors denote different tasks.}
\label{fig:full-dynamic-evaluation}
\end{figure*}

\subsection{Multi-task learning for caricature recognition}
In this section, we evaluate the different approaches on the datasets CaVI and WebCaricature.  ~\figurename~\ref{fig:full-dynamic-evaluation} demonstrates the comparison of the proposed dynamic multi-task learning and the naive multi-task learning~\cite{yin2018multi} for the caricature recognition, visual recognition and the caricature-visual face verification on the two datasets. We can see that the propose method can dynamically update the weights and focus on the training of the  hardest task. For example, in the (a) of \figurename~\ref{fig:full-dynamic-evaluation}, when the loss of the caricature recognition task (denoted by the grey curve in the bottom row of (a)) is high meaning that this is a difficult task, the weight of the caricature recognition task (grey curve in the upper row of (a)) is assigned by a large value to drive the networks to train primarily this task. Thanks to the large weight of task, the loss of the caricature recognition task descends quickly, accordingly the weight of the caricature recognition task begin to descend. Gradually the loss of the caricature recognition task even is lower than the caricature-visual verification task (the blue curve), then at this moment the weight of the caricature recognition task is not the largest anymore and the networks begin to augment the weight of the caricature-visual verification task (the blue curve) to lead the networks to focus on training this new task. However,  (b) of \figurename~\ref{fig:full-dynamic-evaluation} shows that the naive dynamic method always assigns the hard caricature recognition task with a high loss (denoted by the grey curve of the bottom row of (b)) a small weight (denoted by the grey curve of the upper row of (b)). The hard task that is mistakenly assigned the weight cannot be sufficiently trained to reduce the loss so that its weight of task continues to be small. This dilemma hinders the networks to train the tasks efficiently and affect the improvements of the performance. The same situation can be found also on the dataset WebCaricature shown in (c) and (d) of \figurename~\ref{fig:full-dynamic-evaluation}.

Table~\ref{tab:multitaskevalCaVI} shows the evaluation results of the caricature-visual face verification, caricature identification and visual face identification on dataset CaVI. It shows that for all three tasks, the proposed dynamic multi-task learning method outperforms the state-of-art method CaVINet. We also evaluate the naive dynamic multi-task learning method based on our networks. We can see that for the hard task caricature identification, the performance of the naive dynamic multi-task learning (75.80\%) is inferior to our method (85.61\%) and also worse than the performance of the single-task model (78.20\%). Comparing to naive dynamic method, our dynamic multi-task learning gain great improvement of the performance on the caricature identification task (9.81\%). This proves that our dynamic multi-task learning is capable to well train the hard task. And even for the easy tasks our dynamic method can still achieve or improve slightly the performance comparing to the naive dynamic method, i.e. 1.12\% for the caricature-visual verification task and 0.75\% for the visual identification task.  
Table~\ref{tab:WebCariVerif}, ~Table~\ref{tab:WebCariC2P} and ~Table~\ref{tab:WebCariP2C} demonstrate the evaluation results on the dataset WebCaricature.  Since the methods proposed in ~\cite{HuoBMVC2018WebCaricature} are the baseline methods for demonstrating the benchmark WebCaricature, the performance of our methods boost significant comparing to the baseline approaches. All the evaluation are conducted by the 10-folds cross validation by following the evaluation protocol of WebCaricature. We can see that on all tasks, our method has achieve the best performance. However, there is still much room to improve in terms of the weak performance of the validation rate (recall rate) at a low false accept rate (false positive rate).
\section{Conclusion}
 In this work, we propose a multi-task learning approach with dynamic weights for the cross-modal caricature-visual face recognition, which can model the different recognition modalities by the different tasks. The proposed dynamic weight module without introducing the additional hyperparameters can lead the multi-task learning to train the hard task primarily instead of being stuck in the overtraining of the easy task. Both the theoretical analysis and the experimental results demonstrate the effectiveness of the proposed approach to self-adapt the dynamic weights according to the loss of the tasks which also gains a good capacity for cross-modal caricature recognition. Although this multi-task learning approach is proposed for the multi-modalities problems, it can also easily be reproduced on the other multi-task learning frameworks by virtue of the simple structure to generate the dynamic weights.
 \begin{table}
\caption{\label{tab:multitaskevalCaVI}The evaluation of caricature-visual face verification (accuracy\%) on dataset CaVI.}
\begin{center}
\small
\begin{tabular}{|l|c|c|c|}
\hline
Method & Cari-Visual & Visual-id & Cari-id\\
\hline\hline
CaVINet &91.06&94.50&85.09 \\
CaVINet(TW) &84.32&85.16&86.02  \\
CaVINet(w/o) &86.01&93.46&80.43  \\
CaVINet(shared) &88.59&90.56&81.23  \\
CaVINet(visual) &88.58&92.16&83.36  \\\hline
{Naive Dynamic}  &93.80&97.60&75.80  \\
{Ours (verif)}  &92.46&-&-  \\
{Ours (visual)}  &-&98.10&-  \\
{Ours (cari)}  &-&-&78.20  \\
{Ours (Multi-task)}  &\textbf{94.92}&\textbf{98.35}&\textbf{85.61} \\

\hline
\end{tabular}
\end{center}
\end{table}
 \begin{table}
\caption{\label{tab:WebCariVerif}The evaluation of caricature-visual face verification in terms of the validation rate (\%) on dataset WebCaricature.}
\begin{center}
\small
\begin{tabular}{|l|c|c|}
\hline
Method & VAL@FAR=0.1\% & VAL@FAR=1\%\\
\hline\hline
SIFT-Land-ITML & 5.08$\pm$1.82  & 18.07$\pm$4.72  \\
VGG-Eye-PCA &21.42$\pm$2.02&40.28$\pm$2.91  \\
VGG-Eye-ITML &18.97$\pm$3.90&41.72$\pm$5.83  \\
VGG-Box-PCA &28.42$\pm$2.04&55.53$\pm$2.76  \\
VGG-Box &34.94$\pm$5.06&57.22$\pm$6.50\\\hline
{naive Dynamic}  &38.39$\pm$4.58&79.69$\pm$1.3 \\
{Ours (Single-verif)}  &42.10$\pm$3.05&\textbf{84.52$\pm$0.80}\\
{Ours (Multi-task)}  &\textbf{45.82$\pm$1.65}&83.20$\pm$2.00 \\

\hline
\end{tabular}
\end{center}
\end{table}
 \begin{table}
\caption{\label{tab:WebCariC2P} The evaluation of Caricature to Photo identification (C2P) on dataset WebCaricature.}
\begin{center}
\small
\begin{tabular}{|l|c|c|}
\hline
Method & Rank-1(\%) &Rank-10(\%)\\
\hline\hline
SIFT-Land-KCSR & 24.87 $\pm$ 1.50  & 61.57 $\pm$  1.37   \\
VGG-Eye-PCA & 35.07 $\pm$ 1.84  & 71.64 $\pm$  1.32   \\
VGG-Eye-KCSR & 39.76 $\pm$ 1.60  & 75.38 $\pm$  1.34   \\
VGG-Box-PCA & 49.89 $\pm$ 1.97  & 84.21 $\pm$  1.08   \\
VGG-Box-KCSR & 55.41 $\pm$ 1.41  & 87.00 $\pm$  0.92   \\\hline
{naive Dynamic}  & 86.00 $\pm$ 1.70  & 98.21 $\pm$  1.08   \\
{Ours (Single-verif)}  & 85.55 $\pm$ 1.30  & 96.31 $\pm$  0.08   \\
{Ours (Multi-task)}  & \textbf{87.30 $\pm$ 1.20}  & \textbf{99.21 $\pm$  1.07}   \\

\hline
\end{tabular}
\end{center}

\end{table}

 \begin{table}
\caption{\label{tab:WebCariP2C}The evaluation of Photo to Caricature (P2C) identification (C2P) on dataset WebCaricature.}
\begin{center}
\small
\begin{tabular}{|l|c|c|}
\hline
Method & Rank-1(\%) &Rank-10(\%)\\
\hline\hline
SIFT-Land-KCSR & 23.42 $\pm$ 1.57  & 69.95 $\pm$  2.34   \\
VGG-Eye-PCA & 36.18 $\pm$ 3.24  & 68.95 $\pm$  3.25   \\
VGG-Eye-KCSR & 40.67 $\pm$ 3.61  & 75.77 $\pm$  2.63   \\
VGG-Box-PCA & 50.59 $\pm$ 2.37  & 82.15 $\pm$  1.31   \\
VGG-Box-KCSR & 55.53 $\pm$ 2.17  & 86.86 $\pm$  1.42   \\\hline
{naive Dynamic}  & 82.80 $\pm$ 1.60  & 97.81 $\pm$  0.88   \\
{Ours (Single-verif)}  & 81.70 $\pm$ 2.60  & 95.25 $\pm$  1.08   \\
{Ours (Multi-task)}  & \textbf{84.00 $\pm$ 1.60}  & \textbf{99.01 $\pm$  1.2}   \\

\hline
\end{tabular}
\end{center}

\end{table}



{\small
\newcommand{\BIBdecl}{\setlength{\itemsep}{0.25 em}}
\bibliographystyle{IEEEtran}
\bibliography{egbib_final}
}
\end{document}